\documentclass[runningheads]{llncs}

\usepackage{eccv}

\usepackage{eccvabbrv}

\usepackage{graphicx}
\usepackage{booktabs}

\usepackage[accsupp]{axessibility}  

\usepackage[pagebackref,breaklinks,colorlinks,citecolor=eccvblue]{hyperref}
\usepackage{hyperref}

\usepackage{orcidlink}
\usepackage{multirow}

\begin{document}

\title{3D Weakly Supervised Semantic Segmentation with 2D Vision-Language Guidance} 

\titlerunning{3DSS with 2D Vision-Language Guidance}

\author{Xiaoxu Xu\inst{1} \and
Yitian Yuan\inst{2}\orcidlink{0000-0001-8701-7689} \and
Jinlong Li\inst{3}\orcidlink{0000-0002-8746-4566} \and
Qiudan Zhang\inst{1}\orcidlink{0000-0001-6067-8188} \and
Zequn Jie\inst{2}\orcidlink{0000-0002-3038-5891} \and
Lin Ma\inst{2}\orcidlink{0000-0002-7331-6132} \and
Hao Tang\inst{4,5}\orcidlink{0000-0002-2077-1246} \and
Nicu Sebe\inst{3}\orcidlink{0000-0002-6597-7248} \and
Xu Wang\inst{1}\thanks{Corresponding author: wangxu@szu.edu.cn}\orcidlink{0000-0002-2948-6468}
}

\authorrunning{X. Xu et al.}

\institute{College of Computer Science and Software Engineering, Shenzhen University, Shenzhen, 518060, China. \and Meituan Inc., China.\and University of Trento, Italy. \and Peking University, China. \and Carnegie Mellon University, USA.
}

\maketitle

\begin{abstract}
  In this paper, we propose \textbf{3DSS-VLG},  a weakly supervised approach for \textbf{3D} \textbf{S}emantic \textbf{S}egmentation with 2D \textbf{V}ision-\textbf{L}anguage \textbf{G}uidance, an alternative approach that a 3D model predicts dense-embedding for each point which is co-embedded with both the aligned image and text spaces from the 2D vision-language model. Specifically, our method exploits the superior generalization ability of the 2D vision-language models and proposes the Embeddings Soft-Guidance Stage to utilize it to implicitly align 3D embeddings and text embeddings. Moreover, we introduce the Embeddings Specialization Stage to purify the feature representation with the help of a given scene-level label, specifying a better feature supervised by the corresponding text embedding. Thus, the 3D model is able to gain informative supervisions both from the image embedding and text embedding, leading to competitive segmentation performances. To the best of our knowledge, this is the first work to investigate 3D weakly supervised semantic segmentation by using the textual semantic information of text category labels. Moreover, with extensive quantitative and qualitative experiments, we present that our 3DSS-VLG is able not only to achieve the state-of-the-art performance on both S3DIS and ScanNet datasets, but also to maintain strong generalization capability. The code will be available at https://github.com/xuxiaoxxxx/3DSS-VLG/.
\end{abstract}

\section{Introduction}
3D point cloud semantic segmentation~\cite{qi2017pointnet,qi2017pointnet++,hu2020randla,yan20222dpass,qian2022pointnext,hegde2023clip} can provide valuable geometric and semantic data about the 3D environment and has gained considerable attention over the past few years. Learning-based semantic segmentation methods have achieved remarkable performance recently, but they need per-point annotations, which is time consuming and labor intensive.

\begin{figure}[t]
  \centering
   \includegraphics[width=0.89\linewidth]{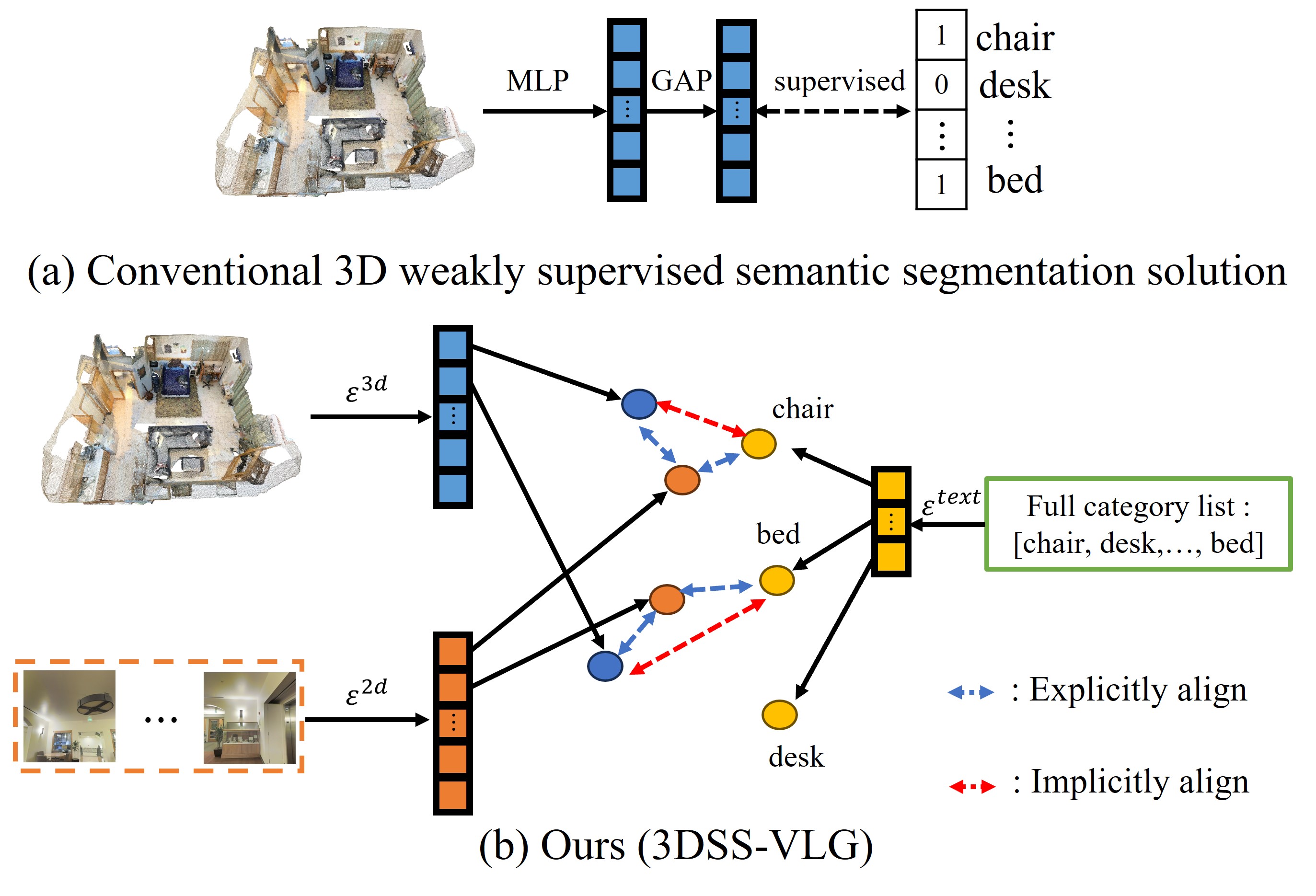}
      \caption{Comparison of different approaches. (a) The conventional 3D WSSS approach adopts the coarse-grained CAM method in a global manner and is supervised by scene-level annotations or subcloud-level annotations. (b) Our proposed 3DSS-VLG approach leverages natural 3D-2D correspondence from geometric camera calibration and 2D-text correspondence from vision-language models, to implicitly align texts and 3D point clouds.}
   \label{fig1}
\end{figure}

To address this issue, existing weakly supervised methods derive the segmentation model with different weak supervisory signals, such as subcloud-level annotations~\cite{wei2020multi}, scene-level annotations~\cite{ren20213d,yang2022mil,kweon2022joint} and so on. As shown in \cref{fig1}~(a), the 3D Weakly Supervised Semantic Segmentation (3D WSSS) approaches typically adopt a Class Activation Map (CAM)~\cite{zhou2016learning} solution. Point clouds are first processed by several Multi-Layer Perception (MLP) layers and thus get a point cloud feature map, and then this point cloud feature map is processed by a Global Average Pooling~(GAP) to get a global classification prediction, which is trained with subcloud-level or scene-level annotations. Given the simple GAP connectivity structure, these methods can easily identify the importance of each point by projecting back the output classification weight onto the point cloud feature maps, a technique we call CAM.  In this way, the semantic segmentation for each category is back-derived from the global prediction. Recently, with the remarkable success of 2D vision, some methods~\cite{kweon2022joint,yang20232d} also use the 2D module to enhance the 3D WSSS.

Although leveraging 2D-3D fusion in 3D WSSS seems to be promising, there also exist some problems. Kweon~\textit{et al.}~\cite{kweon2022joint} need extra detailed annotations of 2D images. As for MIT~\cite{yang20232d}, although it avoids additional per-point/pixel annotations or per-image class labels, its performance is not expected. Therefore, how to design a network that achieves good performance despite the lack of 2D annotations still remains a big challenge. Meanwhile, we notice that the conventional methods for 3D WSSS only use the scene-labels or subcloud-labels to supervise the model, but ignore that the textual category labels such as ``chair, table'' also have semantic meanings and could be embedded to help the model learning. At the same time,  we also find that some methods~\cite{ghiasi2022scaling,liang2023open,yun2023ifseg} like Openseg~\cite{ghiasi2022scaling}, which leverage the pretrained vision-language models such as CLIP~\cite{radford2021learning} to establish precise semantic matching relationships between natural languages and 2D images,  have achieved good results in 2D open vocabulary semantic segmentation (2D OVSS). The above two points inspire us to consider whether we can use the well-pretrained 2D OVSS model to help the 3D WSSS. As shown in \cref{fig1} (b), the point cloud and 2D images could be mutually mapped with geometric projections, and the 2D images and textual categories could be compared with pretrained vision-language models. Therefore, why do not we take the 2D images as a bridge, leveraging the correspondences between point clouds and images, images and natural languages, to implicitly build matching relationships between point clouds and natural languages?

To this end, we propose a simple yet effective method, namely \textbf{3DSS-VLG}, \textit{i.e.,} a weakly supervised approach for \textbf{3D} \textbf{S}emantic \textbf{S}egmentation with 2D \textbf{V}ision-\textbf{L}anguage \textbf{G}uidance. Our 3DSS-VLG only needs to use 2D images, but no need for their 2D image-level annotations during training. Specifically, for the input 3D point cloud, the dataset also provides a set of multi-view images corresponding to it. We first process these multi-view images using the image encoder of the pretrained off-the-shelf 2D OVSS model such as Openseg~\cite{ghiasi2022scaling} to get the 2D embeddings. Then, for each point in the 3D point cloud, we project it to the multi-view images with geometric projections, and integrate these corresponding 2D embeddings to get the 2D-projected embeddings for the point. Next, we utilize the text module of the 2D OVSS model to obtain the textual embeddings of each semantic category label. Since in the embedding space of 2D OVSS, the textual category labels and 2D images could be directly compared, we only need to learn a 3D backbone which could generate 3D embeddings aligned with 2D embeddings; thus, the category labels and the 3D point cl
oud could be implicitly compared. 

Actually, if the 3D embedding is learned well enough, it can be directly compared with the text embedding by the similarity measurement to classify. However, we find that only relying on pulling the 2D-projected embeddings and 3D embeddings closely is not reliable since the pretrained 2D OVSS model are designed to learn the general knowledge and do not have specialized knowledge to the indoor point cloud scene. Therefore, we propose to alleviate this problem by three stages. (1) First, as shown in \cref{fig2},  we perform matrix multiplication on projected 2D embeddings and text embeddings of category labels and get the classification logits. Then, we use the scene-level labels as mask to filter out some confusing and unreliable predictions in the classification logits and thus get a more reliable pseudo label vector. (2) Second, as shown in \cref{fig3}~(a), we propose the Embeddings Specialization Stage, which transfers the 2D-projected embeddings with an adapter module to obtain adapted 3D embeddings, and the training of this adapter module will be supervised with the pseudo label vector. This stage is designed to induce a more reliable target 3D embeddings suited for the indoor point cloud scene from the 2D-projected embeddings. (3) Finally, as shown in \cref{fig3}~(b), we design Embeddings Soft-Guidance Stage, which freezes the adapter module introduced in the second stage and leverages cosine similarity to align the adapted 3D embeddings and the MinkowskiNet~\cite{choy20194d} 3D embeddings. Combining the above three stages, we can learn a more reliable 3D embedding space for semantic segmentation in indoor point cloud scene. In the inference procedure, we only need to compare the MinkowskiNet 3D embeddings of the point cloud and the text embeddings of the semantic category labels, thus accomplishing the 3D semantic segmentation. Note that we do not need 2D images to participate in the inference process of our model. 

In summary, the main contributions of this paper are as follows:
\begin{itemize}
    \item  We propose a weakly supervised method 3DSS-VLG for 3D WSSS, which takes 2D images as a bridge, and leverages natural 3D-2D correspondence from geometric camera calibration and 2D-text correspondence from vision-language models to implicitly establish the semantic relationships between texts and 3D point clouds.
    \item We design a three-stage training procedure to learn a reliable 3D embedding space in 3DSS-VLG for 3D semantic segmentation. Pseudo Label Generation Stage is designed to utilize the pretrained 2D vision-language model to provide a embedding space for 3D point cloud representation with MinkowskiNet 3D backbone. Moreover, we propose Embeddings Specialization Stage to make the embedding space to be more robust based on the pseudo label filtering with indoor point cloud scene knowledge.
    \item Extensive experiments on the ScanNet and S3DIS dataset show that the proposed 3DSS-VLG significantly outperforms the previous state-of-the-art methods, even Kweon \textit{et al}. which use extra 2D image-level annotations. Moreover, our futher experiments show our 3DSS-VLG has strong generalization capability and can be extended to handle unobserved general datasets.
\end{itemize}

\section{Related Work}
\subsection{2D Open-Vocabulary Semantic Segmentation}
The recent advances of large vision-language models have enabled a remarkable level of robustness in open-vocabulary semantic segmentation~\cite{liang2023open,xu2022simple,xu2023side,xu2023learning,chen2023exploring}. Open vocabulary semantic segmentation aims to segment the target categories that cannot be access during the training procedure. Pioneering work ZS3Net\cite{bucher2019zero} uses generative models to synthesize pixel-level features by word embeddings of unseen classes. SPNet~\cite{xian2019semantic} encodes visual features into the semantic embeddings space to align with text embeddings. More recently, researchers propose to leverage the pretrained CLIP\cite{radford2021learning} for open-vocabulary semantic segmentation. ZSSeg\cite{xu2022simple} leverages the visual module to generate class-agnostic masks and uses the pretrained text encoder to retrieve the unseen class masks. OpenSeg\cite{ghiasi2022scaling} proposes to align the segment-level visual embeddings with text embeddings via region-word grounding. In this work, we solely rely on pretrained 2D open-vocabulary models and perform 3D weakly supervised semantic segmentation understarnding tasks. We pull the 3D embeddings and 2D embeddings which features exacted from pretrained model back-project onto point cloud closed to implicitly align 3D embeddings and text embeddings. 

\subsection{3D Weakly Supervised Semantic Segmentation}
This task aims to learn point cloud semantic segmentation using weakly annotated data, such as sparsely labeled points~\cite{hu2022sqn,zhang2022not}, box-level labels~\cite{chibane2022box2mask}, subcloud-level labels~\cite{wei2020multi} and scene-level labels~\cite{ren20213d,kweon2022joint,yang20232d,choy20194d}. Though the state-of-the-art methods based on sparsely labeled points show performance comparable to that of supervised ones, they require at least partial point-wise annotation in a scene, which is still expensive compared to subcloud-level labels and scene-level labels. The pipeline of the conventional CAM solution has been used in the majority of previous 3D WSSS works and only treats the scene-level labels as one-hot digit. MPRM\cite{wei2020multi} proposes the subcloud-level annotations method that samples subclouds from the full scene and annotates them, which can alleviate the class imbalance issue commonly appearing in almost scene. However, the subcloud-level annotations need to divide the point cloud into small that we need to annotations more than one for a scene, which is too much trouble and time-consuming. Therefore, some methods that use scene-level annotations are proposals for the 3D WSSS. Kweon \textit{et al.}~\cite{kweon2022joint} utilizes 2D and 3D data for semantic segmentation and gets good performance, however, requiring extra 2D image-level labels. MIT~\cite{yang20232d} proposes the interlaced transformer structure to fuse 2D-3D information with only scene-level labels. However, its performance is not as good as expected.

Therefore, in this work, we explore a 3D WSSS method with only scene-level labels.  Unlike those previous works, we use the semantic meanings of textual category labels to assist in model learning. Moreover, the performance of our 3DSS-VLG is over the Kweon \textit{et al.}, which uses extra 2D image-level labels.

\subsection{2D Semantic in 3D task}
Studies on 3D object detection and semantic segmentation~\cite{genova2021learning,alonso20203d,li2022expansion,cardace2023exploiting,ando2023rangevit,li2024enhancingrobustnessvisionlanguagemodels,li2023mseg3d,li2022weakly,hou2021pri3d,li2023weakly} have explored the use of 2D image semantics to assist 3D tasks. There are almost two approaches: concatenating the image embeddings with each point in the 3D scene as extra information~\cite{hu2021bidirectional,xu2023weakly,zhang2023learning,chen2024towards} or projecting image semantic results into a 3D space to assist 3D semantic segmentation~\cite{xu2018pointfusion,lahoud20172d,qi2018frustum}. Previous studies usually used 2D image semantics as extra inputs in both training and inference. Although performance has improved, the extra 2D inputs have the potential to constrain the range of application scenarios. This is due to the fact that 2D information may be absent during inference or necessitate laborious pre-processing. In this paper, we aim to investigate the potential of using 2D semantics exclusively during training to assist in the 3D WSSS task.

\section{The Proposed Method}
\begin{figure*}[t]
\centering
\includegraphics[width=0.92\linewidth]{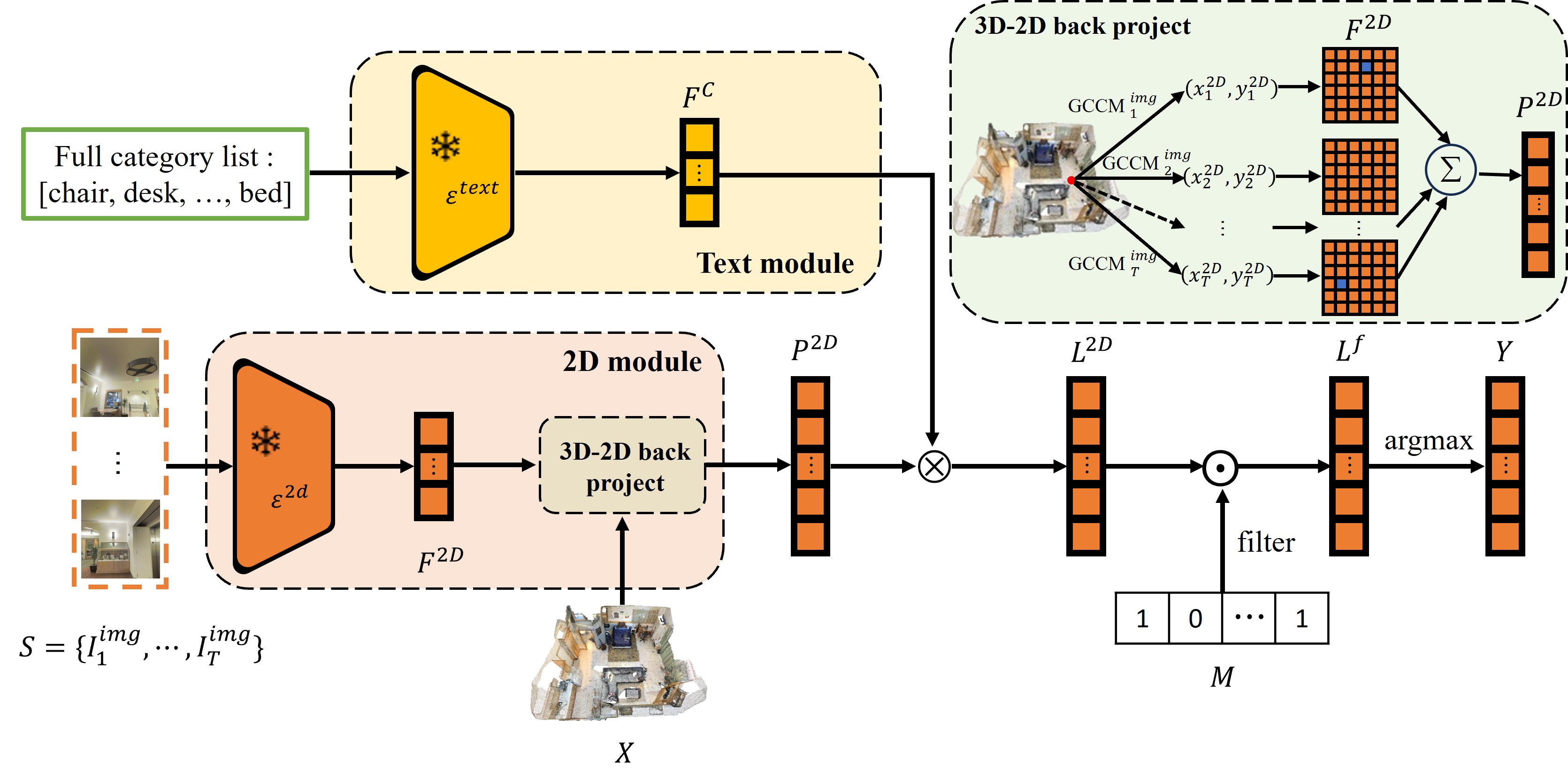}
\caption{\small The proposed pseudo label generation procedure. We first leverage the text encoder $\varepsilon^{text}$ of Openseg to get embeddings of the full category labels $\mathbf{F}^C$, and leverage the 2D image encoder $\varepsilon^{2D}$ of Openseg to get embeddings of the 2D image $\mathbf{F}^{2D}$. It is important to note that we freeze the whole Openseg model during the procedure of pseudo label generation. Then we back-project the 2D embeddings $\mathbf{F}^{2D}$ to integrate the 2D-projected embeddings $\mathbf{P}^{2D}$. Specifically, for each point in the point cloud~$(x^{3D}, y^{3D},z^{3D})$, we use geometric camera calibration matrixes $GCCM^{img}$ to calculate the corresponding positions~$(x^{2D}, y^{2D})$ on the multi-view images $S$. Then we integrate these corresponding 2D embeddings in $\mathbf{F}^{2D}$ and average them to get the 2D-projected embeddings $\mathbf{P}^{2D}$.  We perform matrix multiplication on $\mathbf{F}^{C}$ and $\mathbf{P}^{2D}$, and get the 3D point cloud semantic segmentation prediction logits $\mathbf{L}^{2D}$. Finally we utilize the scene-level labels as mask $M$ to filter out some confusing and unreliable predictions in the classification and get the more accurate predicted logits $\mathbf{L}^{f}$ and pseudo labels~$\mathbf{Y}$.}
\label{fig2}
\end{figure*}
In this section, we will first introduce the procedure of pseudo label generation in \cref{sec3.1}. Then, we will demonstrate the training procedure of our 3DSS-VLG in \cref{sec3.2} and \cref{sec3.3}. Finally, we will
describe the 3DSS-VLG inference procedure in \cref{sec3.4}. 

\subsection{Pseudo Label Generation Stage}
\label{sec3.1}
This stage aims to utilize the pretrained vision-language model and scene-level labels to generate more precise pseudo label. Given an input point cloud with multi-view images as shown in \cref{fig2}, we first implement dense 2D embeddings extraction for each RGB image via the frozen visual encoder of Openseg~\cite{ghiasi2022scaling}, and back-project them onto the 3D surface points of a scene to integrate the 2D-projected embeddings. Afterward, more accurate pseudo labels are generated based on 2D-projected embeddings, text embeddings and scene-level labels.

\noindent\textbf{2D Embeddings Extraction.}
The inputs of 3DSS-VLG comprise a scene with 3D point cloud, scene-level labels and the associated multi-view RGB images set. Given the RGB images set $S$ consists of $T$ images with a resolution of $H \times W$. The point cloud $\mathbf{X} \in\mathbb{R}^{N\times6}$ contains $N$ points in the scene, and each point is represented with six dimensions of RGBXYZ. We leverage the pretrained image encoder of OpenSeg\cite{ghiasi2022scaling} to get per-pixel embedding, denoted as $\mathbf{F}^{2D} \in\mathbb{R}^{T\times H\times W\times d}$, where $d$ is the 2D embedding dimension. For each point in the 3D point cloud, we project it onto multi-view images through geometric camera calibration matrixes and get the corresponding 2D positions. Then we can exact the corresponding projected 2D embeddings from $\mathbf{F}^{2D}$ according to the calculated 2D image positions. Since each point may have multiple correspondences in different images, the final 2D-projected embeddings $\mathbf{P}^{2D}\in\mathbb{R}^{N\times d}$ is obtained via average all the corresponding projected 2D embeddings of each point. 

\noindent\textbf{Text Embeddings Extraction.}
We take the text encoder of Openseg to exact text embeddings $\mathbf{F}^{C}\in\mathbb{R}^{K\times d}$ of full category labels, where $K$ denoted the number of categories. Similarly, we also freeze the text encoder and directly load the pretrained Openseg parameters. 

\noindent\textbf{Filtering Strategy.}
After getting the 2D-projected embeddings $\mathbf{P}^{2D}$ and the text embeddings $\mathbf{F}^{C}$, we perform matrix multiplication on them and obtain the classification logits $\mathbf{L}^{2D}\in\mathbb{R}^{N\times K}$. To make classification logits more reliable, the filtering strategy is employed to filter out confusing and unreliable predictions. For instance, as shown in~\cref{fig2}, we create a boolean scene-level label mask $\mathbf{M}\in\mathbb{R}^{1\times K}$, where the element value in the mask indicated whether the corresponding category existed. Finally, we perform matrix inner product on classification logits $\mathbf{L}^{2D}$ and scene-level label mask $\mathbf{M}$ and obtain filtered classification logits $\mathbf{L}^f \in\mathbb{R}^{N\times K}$. After ranking the filtered classification logits $\mathbf{L}^f$, we can get the more precise pseudo label $\mathbf{Y} \in\mathbb{R}^{N} $ of the input point cloud.

\begin{figure}[t]
\centering
\includegraphics[width=0.92\linewidth]{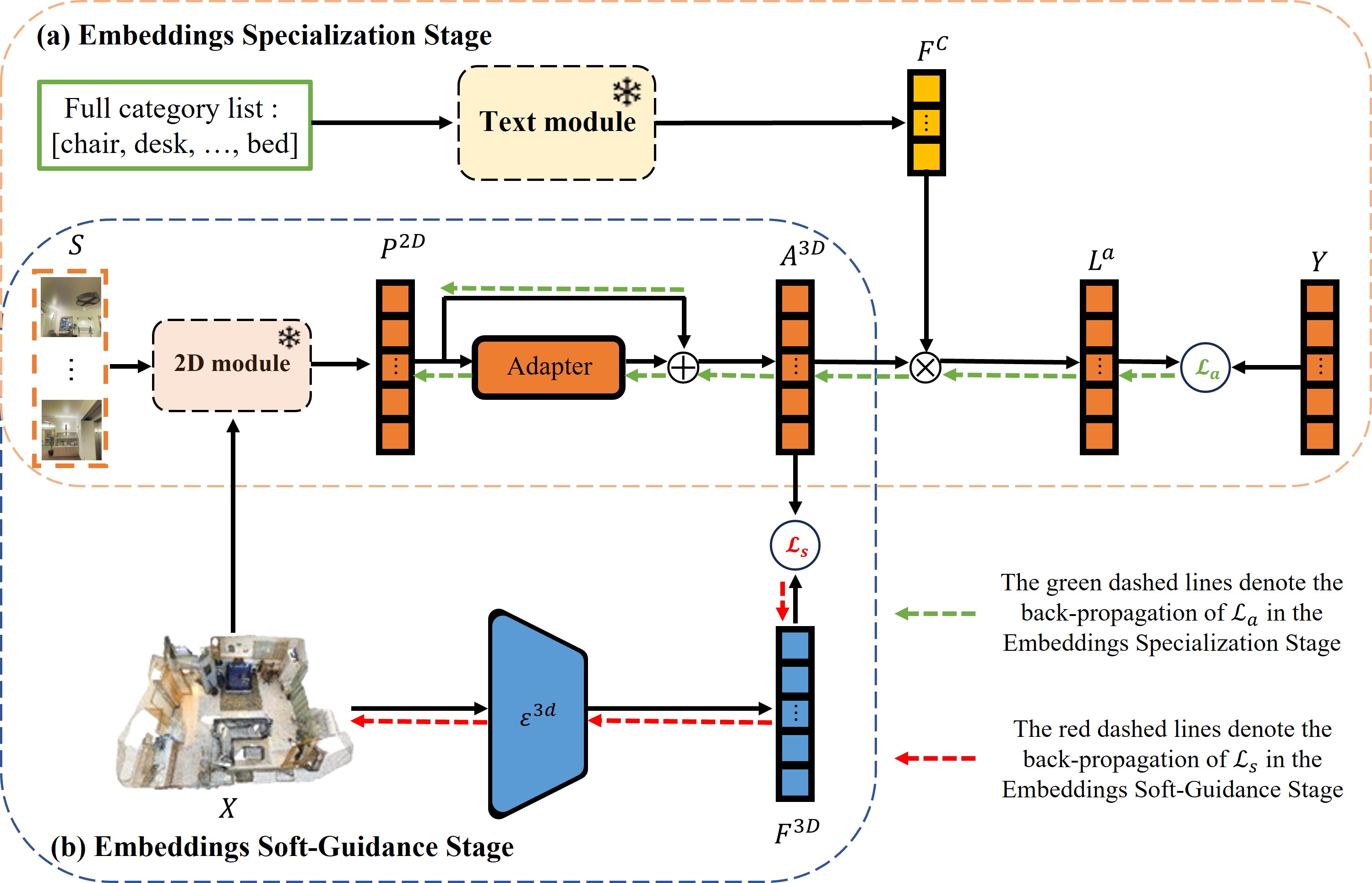} 
\caption{\small The proposed training procedure of our proposed 3DSS-VLG. Here, it is mainly divided into two stages: (a) Embeddings Specialization Stage and (b) Embeddings Soft-Guidance Stage. For (a), we first utilize the text encoder $\varepsilon^{text}$ of Openseg to obtain embeddings of the category labels $\mathbf{F}^C$, which are frozen during the training procedure of (a). Meanwhile, we get the initial 2D-projected embeddings $\mathbf{P}^{2D}$ from the 2D module and leverage the adapter module to transfer the $\mathbf{P}^{2D}$ to a new embedding spaces to obtain the adapted 3D embeddings $\mathbf{A}^{3D}$. We perform matrix multiplication on $\mathbf{A}^{3D}$ and $\mathbf{F}^C$ and get the predicted probability $\mathbf{L}^{a}$. Finally, we use the pseudo labels $\mathbf{Y}$ to supervise the model, and the green dashed lines denote back-propagation of the loss $\mathcal{L}_a$. For (b), we first utilize the adapter module and obtain the adapted 3D embeddings $\mathbf{A}^{3D}$. It is important to note that we freeze the adapter module during the training procedure of (b). Meanwhile, we use the 3D module $\varepsilon^{3D}$ to obtain the 3D embeddings $\mathbf{F}^{3D}$. The cosine similarity loss $\mathcal{L}_s$ will be integrated to train the model. The red dashed lines denote back-propagation of the loss $\mathcal{L}_s$.} 
\label{fig3}
\end{figure}

\subsection{Embeddings Specialization Stage}
\label{sec3.2}
As we know, the 2D OVSS model is designed to learn general knowledge and do not have any specialized knowledge of the indoor point cloud scene. Therefore, only relying on the 2D embeddings to build the 3D-text correlation will make the 3D WSSS process not reliable. To mitigate this issue, the Embeddings Specialization Stage is proposed to further improve the perception of indoor knowledge of 3D embeddings.

Specifically, the 2D-projected embeddings $\mathbf{P}^{2D}$ of input are transferred into another space through the adapter module, which simply contains two fully-connected layers. Besides, to keep both the source and adapted semantics, we employ the residual connections to get the adapted 3D embeddings $\mathbf{A}^{3D}\in\mathbb{R}^{N\times d}$:
\begin{equation}
\mathbf{A}^{3D}=\alpha\cdot MLP(\mathbf{P}^{2D})+(1-\alpha)\cdot \mathbf{P}^{2D},
\end{equation}
where $\alpha$ is the ratio of residual connections. Next, we perform matrix multiplication on text embeddings $\mathbf{F}^{C}$ and adapted 3D embeddings $\mathbf{A}^{3D}$ and obtain the classification logits $\mathbf{L}^{a}\in\mathbb{R}^{N\times K}$. The softmax layer is applied on $\mathbf{L}^{a}$ and a classification cross-entropy loss $\mathcal{L}_{a}$ is introduced to supervise the procedure. Here we leverage the pseudo labels $\mathbf{Y}$ of the point cloud to supervise the model.

Just by introducing the above simple adapter module, we can make the learned adapted embeddings have better semantic awareness of the point clouds of indoor scenes, thus assisting the 3D WSSS task.

\subsection{Embeddings Soft-Guidance Stage}
\label{sec3.3}
Since Openseg has established a high level of semantic alignment between 2D embeddings and text embeddings, we propose the Embeddings Soft-Guidance Stage, which can naturally take the 2D embeddings as a bridge to implicitly align the 3D embeddings and text embeddings via cosine similarity. Specifically, as shown in \cref{fig3} (b), we take the point cloud $\mathbf{X}$ as input, and use MinkowskiNet18A UNet\cite{choy20194d} as our 3D module meanwhile, we change the dimension of the outputs to $d$. Therefore, we can get the learned 3D embeddings $\mathbf{F}^{3D}\in\mathbb{R}^{N\times d}$. Then we take the corresponding 2D-projected embeddings $\mathbf{P}^{2D}$ as input, processed by the adapter module, and get the adapted 3D embeddings $\mathbf{A}^{3D}$. We follow the typical cosine similarity loss by pulling the paired 3D embeddings $\mathbf{F}^{3D}$ and adapted 3D embeddings $\mathbf{A}^{3D}$ closer. We need to note that we freeze the adapter module and directly load the parameters provided by \cref{sec3.2} during training. 
Therefore, we define the 3DSS-VLG loss as:

\begin{equation}\label{eq1}
\mathcal{L}_s=1-\cos\left(\mathbf{F}^{3D}, \mathbf{A}^{3D}\right).
\end{equation}

\subsection{Inference}
\label{sec3.4}
During inference, we only retain the 3D and text modules and remove the 2D module. Specifically, we take the 3D embeddings $\mathbf{F}^{3D}$ from the 3D module, as well as the category embeddings $\mathbf{F}^{C}$ from text module, to perform matrix multiplication on them and get the classification logits. Finally, we rank the logits and obtain the final per-point segmentation for the input point cloud $\mathbf{X}$.

\section{Experiments}
In this section, we first present our experimental settings, including datasets, evaluation metrics, and implementation details. The competing methods are then presented and compared. Finally, ablation studies are provided to further demonstrate the necessity and effectiveness of each component of our framework.

\subsection{Datasets and Evaluation Metrics}
We evaluate our 3DSS-VLG on two publicly and widely-used large-scale point cloud datasets with multi-view images, S3DIS~\cite{armeni20163d} and ScanNet~\cite{dai2017scannet}. S3DIS is proposed for indoor scene understanding. It consists of 6 areas including 271 rooms with 13 classes. Each room is scanned via RGBD sensors and is represented by a point cloud with 3D coordinates and RGB values. Following previous works, we take area 5 as the test scene. ScanNet~\cite{dai2017scannet} has 1513 training scenes and 100 test scenes with 20 classes. We adopt the default train-val split setting, where there are 1201 training scenes and 312 validation scenes. The mean intersection over Union (mIoU) is employed as the evaluation metric for datasets.

\subsection{Implementations Details}
3DSS-VLG is implemented by PyTorch. For the training procedure of \cref{sec3.2}, we use Adam optimizer with batch size of 16 and set an initial learning rate of 0.003 for the model. We reduce the learning rate by a multiplying factor of 0.7 every 20 epochs for a total of 80 epochs. For the training procedure of \cref{sec3.3}, the model optimization is conducted using Adam optimizer with a batch size of 8. We set an initial learning rate of 0.0001 for the model and use the poly learning rate policy to adjust the learning rate.

\subsection{3D Semantic Segmentation Results}
We evaluate our proposed approach against state-of-the-art techniques for 3D weakly supervised semantic segmentation with scene-level labels. Firstly, we demonstrate some full supervised point cloud semantic segmentation methods to compare the gap between the performances of ours and full supervised methods. Subsequently, we introduce semantic segmentation methods supervised by scene-level labels or subcloud-level labels and compare them with our method. Meanwhile, we indicate the average annotation time per scene.

\begin{table}[t]
\begin{center}
\caption{Performance comparison on the S3DIS dataset. ``Sup.'' indicates the type of supervision. ``100\%'' represents full annotation. ``scene.'' denotes scene-level annotation.}
\begin{tabular}{cccc}
\hline 
Method                                         & Label Effort                         & Sup.                                  & Test \\
\hline 
PointNet~\cite{qi2017pointnet}                 & \multirow{8}{*}{\textgreater 20 min} & 100\%                                 & 41.1 \\
TangentConv~\cite{tatarchenko2018tangent}      &                                      & 100\%                                 & 52.8 \\
MinkowskiNet~\cite{choy20194d}                 &                                      & 100\%                                 & 65.8 \\
KPConv~\cite{thomas2019kpconv}                 &                                      & 100\%                                 & 67.1 \\
PointTransformer~\cite{zhao2021point}          &                                      & 100\%                                 & 70.4 \\
PointNetXt~\cite{qian2022pointnext}            &                                      & 100\%                                 & 70.5 \\
DeepViewAgg~\cite{robert2022learning}          &                                      & 100\%                                 & 67.2 \\
SemAffiNet~\cite{wang2022semaffinet}           &                                      & 100\%                                 & 71.6 \\
\hline 
MPRM~\cite{wei2020multi}                       & \multirow{5}{*}{\textless 1 min}     & scene.                                & 10.3 \\
MIL-Trans~\cite{yang2022mil}                   &                                      & scene.                                & 12.9 \\
WYPR~\cite{ren20213d}                          &                                      & scene.                                & 22.3 \\
MIT~\cite{yang20232d}                          &                                      & scene.                                & 27.7 \\
Ours                                           &                                      & scene.                                & \textbf{45.3} \\
\hline 
\end{tabular}
\label{Tab.1}
    
\end{center}
\end{table}

\noindent\textbf{Evaluation on S3DIS.}
\cref{Tab.1} Show the performance of each type of 3D point cloud semantic segmentation methods evaluated on the S3DIS dataset. We can find that in the scene-level annotations setting, our method greatly surpasses the existing state-of-the-art method MIT~\cite{yang20232d} by 17.6\%. This shows that using textual semantic information ignored by previous 3D weakly supervised semantic segmentation can significantly improve segmentation performance. The textual semantic information of each category is unique; then the 2D embeddings and 3D embeddings are aligned so that the 3D embeddings can be implicitly aligned to the corresponding unique category semantic information, which allows the model to achieve greater performance improvements.

Meanwhile, we compare our method with some full supervised methods. It can be observed that our 3DSS-VLG can outperform some fully supervised methods, \textit{i.e.}, PointNet~\cite{qi2017pointnet}. Moreover, we notice that the annotations cost time of different types of supervision and find that the scene-level annotation is the most efficient compared to other types annotations. Such results demonstrate the effectiveness and potential of our weakly supervised method.

\begin{table}[t]
\begin{center}
\caption{Performance comparison on the ScanNet test set and  validation set. ``Sup.'' indicates the type of supervision. ``100\%'' represents full annotation. ``subcloud.'' and ``scene.'' imply subcloud-level annotation and scene-level annotation respectively. ``image.''  denotes image-level annotation.}
\begin{tabular}{ccccc}
\hline 
Method                                         & Label Effort                                & Sup.     & Test   &Val  \\
\hline 
PointNet++~\cite{qi2017pointnet++}             & \multirow{8}{*}{\textgreater 20 min}        & 100\%    & 33.9   & -\\
TangentConv~\cite{tatarchenko2018tangent}      &                                             & 100\%    & 43.8   & -\\
MinkowskiNet~\cite{choy20194d}                 &                                             & 100\%    & 73.6   & 72.2\\
KPConv~\cite{thomas2019kpconv}                 &                                             & 100\%    & 68.6   & 69.2\\
PointTransformer~\cite{zhao2021point}          &                                             & 100\%    & -      & 70.6\\
PointNetXt~\cite{qian2022pointnext}            &                                             & 100\%    & 71.2   & 71.5\\
DeepViewAgg~\cite{robert2022learning}          &                                             & 100\%    & -      & 71.0 \\
SemAffiNet~\cite{wang2022semaffinet}           &                                             & 100\%    & 74.9   & - \\
\hline 
MPRM~\cite{wei2020multi}                       & 3 min                                      & subcloud. & 41.1   & 43.2 \\
\hline 
Kweon \textit{et al.}~\cite{kweon2022joint}    & 5 min                            & scene. + image.     & 47.4    & 49.6 \\
\hline 
MIL-Trans~\cite{yang2022mil}                   &  \multirow{4}{*}{\textless 1 min}          & scene.    & -       & 26.2 \\
WYPR~\cite{ren20213d}                          &                                            & scene.    & 24.0    & 29.6 \\
MIT~\cite{yang20232d}                          &                                            & scene.    & 31.7     & 35.8 \\
Ours                                           &                                            & scene.    & \textbf{48.9}  & \textbf{49.7} \\
\hline 
\end{tabular}
\label{Tab.2}
\end{center}
\end{table}

\noindent\textbf{Evaluation on ScanNet.} We also evaluate our 3DSS-VLG on the ScanNet online test set and the validation set and presented the performance results of 3DSS-VLG in \cref{Tab.2}. For the test set, it can be observed that our 3DSS-VLG achieves the best performance under only scene-level label supervision and even surpasses the performance of MPRM~\cite{wei2020multi} which is supervised by subcloud-level annotations. Moreover, we are surprised to find that our method also outperforms Kweon \textit{et al.}~\cite{kweon2022joint}  by 1.5\%, which uses not only scene-level labels, but also extra image-level labels. Our method can achieve stronger performance with less annotations, further illustrating the superiority of our method. Meanwhile, our 3DSS-VLG can outperform some fully supervised methods. In addition for the validation set, our method also achieves the state-of-the-art during those 3D WSSS approaches. Those results demonstrate the superiority of 3DSS-VLG.

\subsection{Ablation Studies}
\label{sec:ablation}

\noindent\textbf{Effectiveness of Each Components.}
To demonstrate the advantage of each component in our 3DSS-VLG, we conduct comprehensive ablation studies on the S3DIS dataset, as shown in \cref{Tab.3}. The ablation model (a) only retains the MinkowskiNet18A UNet~\cite{choy20194d} and trains directly with the pseudo labels which are generated without using scene-level labels filtering. The cross-entropy loss is introduced to supervised this procedure. We set model (a) as the baseline of our experiment. Compared to model (a), model (b) is not directly supervised by pseudo labels. It adopts the Embeddings Soft-Guidance Stage (ESGS) and is soft-guided by the 2D-projected embeddings $\mathbf{P}^{2D}$ . We can find that the performance of mIoU is improved from 37.7\% to 38.2\%. This observation proves that the soft-guidance strategy can guide 3D embeddings to align with the text embeddings and achieve better performance compared to directly using the pseudo labels to supervised 3D model. Meanwhile, when we introduce the filtering strategy to model (a), as shown in model (c), we can find that the model performance increases greatly from 37.7\% to 42.6\%. Finally, by adding the filtering strategy to model (b) and utilizing the Embeddings Specialization Stage (ESS), model (d) is supervised by adapted 3D embeddings $\mathbf{A}^{3D}$ at this time. It can be observed the performance improves from 38.2\% to 45.3\%. Such results prove that our 3DSS-VLG can help the model to get a better, indoor point cloud specific embedding space to align 3D point clouds and text.

\begin{table}[t]
\begin{center}
\caption{Ablation studies of the 3DSS-VLG components on S3DIS dataset.}
\begin{tabular}{ccccc}
\hline 
    & ESGS       & Filtering  & ESS        & mIoU \\
\hline 
(a) &            &            &            & 37.7 \\
(b) & \checkmark &            &           & 38.2 \\
(c) &            & \checkmark &            & 42.6 \\
(d) & \checkmark & \checkmark & \checkmark & 45.3 \\
\hline 
\end{tabular}
\label{Tab.3}
\end{center}
\end{table}

\begin{table}[t]
\begin{center}
\caption{Performance comparisons of the generalization capability.}
\begin{tabular}{cccccc}
\hline 
Domain                      & mIoU  & mAcc  \\
\hline 
S3DIS -\textgreater ScanNet & 13.4 & 23.0 \\
ScanNet -\textgreater S3DIS & 33.3 & 50.9 \\
\hline 
\end{tabular}
\label{Tab.4}
\end{center}
\end{table}

\noindent\textbf{Generalization Capability.}
Due to the domain gap among different datasets, a model trained on one dataset is not applicable to another dataset. Also, this situation occurs in the 3D WSSS task. Nevertheless, we notice that, compared to previous works, our 3DSS-VLG uses textual semantic information as a guide rather than CAM, which means our model has a good relationship between 3D point cloud and the text of category labels and indicates that the model may have generalization ability. Therefore, we further explore our framework to the novel data of the unobserved scene domains.

As shown in Tab.~\ref{Tab.4}, we experimentally verify the generalizability of the proposed method on the S3DIS and ScanNet dataset, respectively. The first row is the performance of model that we first train our model on the S3DIS dataset and then test the trained model on validation set of the ScanNet dataset. The second row is the performance of model that we first train our model on the ScanNet dataset and then test the trained model on the test set of the S3DIS dataset. Compared to those weakly supervised methods with scene-level labels, it can be observed that our 3DSS-VLG has a certain gap with those methods in the first row. However, for the second row, we are supervised to find that our method can outperform all the weakly supervised methods and achieve state-of-the-art performance. The ScanNet dataset provides six times more training scenes than the S3DIS dataset. Therefore, when a model is pretrained on the ScanNet dataset, the model will be more robust than a model pretrained on the S3DIS dataset. Our experimental results also prove this phenomenon.

The results also strongly support the complementary advantages of using text semantic information, even without any further fine-tuning or domain-specific adaptation. Our 3DSS-VLG can be extended to handle unobserved general data and has strong generalization capability, which is promising for the field of 3D WSSS.

\begin{table}[t]
\begin{center}
\caption{Performance comparisons with different 3D backbones and ESS module backbones on the S3DIS dataset.}
\begin{tabular}{ccc}
\hline 
Module &Backbone     & mIoU \\
\hline 
\multirow{3}{*}{3D} & MinkowskiNet14A & 44.5 \\
& MinkowskiNet18A & 45.3 \\
& MinkowskiNet34A & 44.7 \\
\hline 
\multirow{2}{*}{ESS} & Transformers & 45.0 \\
& MLP & 45.3 \\
\hline 
\end{tabular}
\label{Tab.5}
\end{center}
\end{table}

\noindent\textbf{Experiments with Different Backbones.}
Tab.~\ref{Tab.5} shows the performances of our method on S3DIS with different 3D backbones and ESS module backbones. Finally, we use the MinkowskiNet18A as our 3D backbone and the FC-layer as the backbone of our ESS.

\begin{figure*}[t]
\centering
\includegraphics[width=1.0\linewidth]{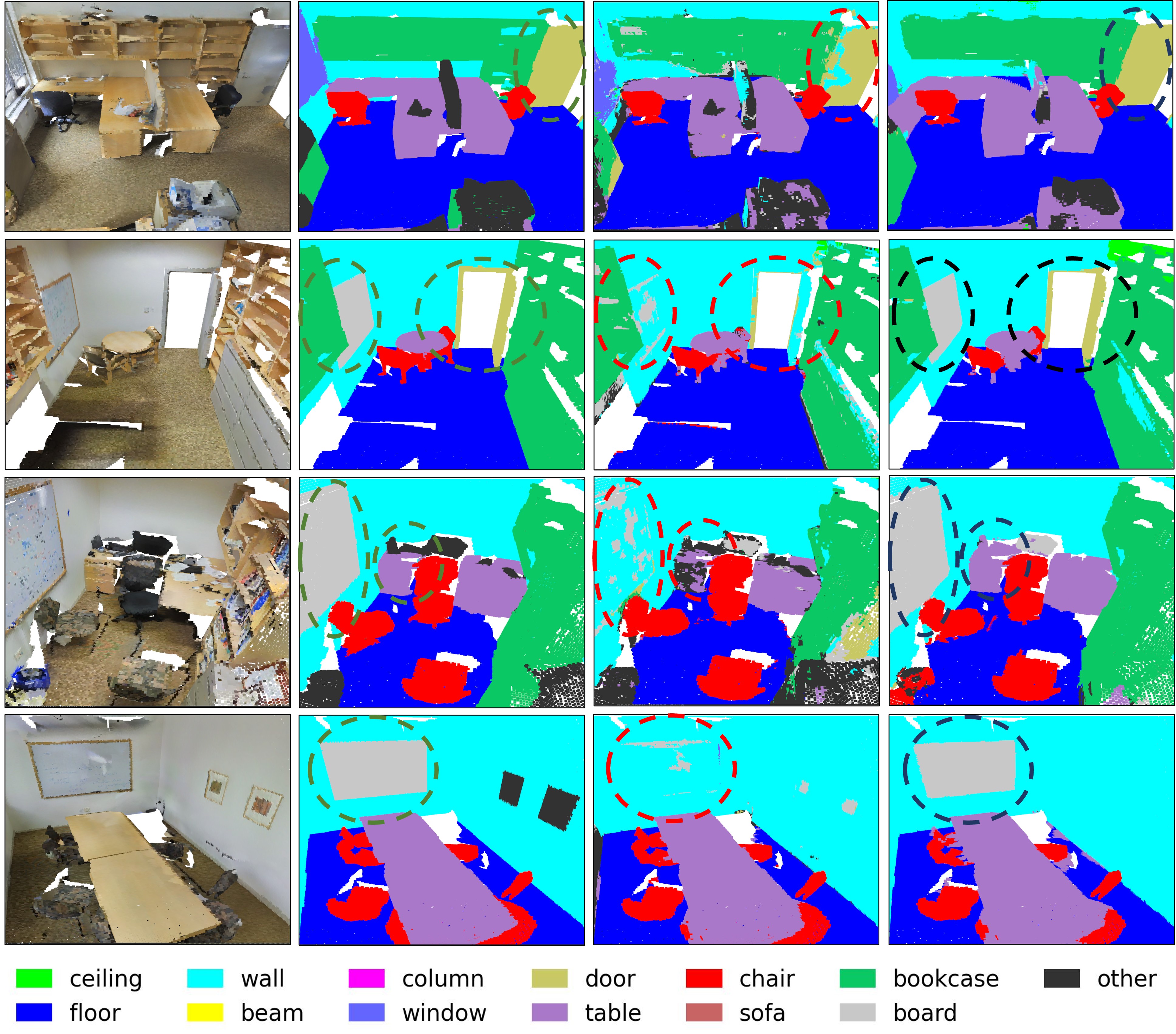} 
\caption{Qualitative results on the S3DIS dataset of baseline and our 3DSS-VLG. From left to right: input point clouds, ground truth, baseline results, and our 3DSS-VLG results.}
\label{fig4}
\end{figure*}

\subsection{Qualitative Results}
\cref{fig4} visualizes the qualitative comparison of the proposed framework and baseline. Here the baseline is model (a) which is mentioned in \cref{sec:ablation}. Compared with the result of baseline, our 3DSS-VLG shows significantly better results in the terms of accuracy of semantics and preciseness of segmentation. With the ESGS, ESS and filtering strategies, our 3DSS-VLG can learn a more better indoor point cloud specific embedding space to align 3D point clouds and text and achieve substantial semantic segmentation results compared to the baseline. 

\subsection{Limitations}
Our work relies on vision-language alignment and does not address how to align visual embeddings with some abstract category text embeddings (\textit{e.g.} ``other'' class in the S3DIS dataset). It is difficult for the model to understand what the difference is between the ``other'' class and other categories, thus making the wrong segmentation. This limitation is a direct avenue for future work.

\section{Conclusion}
In this paper, we propose 3DSS-VLG to address the shortage of point-level annotations. Specifically, our 3DSS-VLG exploits the superior ability of current vision-language models on aligning the semantics between texts and 2D images, as well as the naturally existing correspondences between 2D images and 3D point clouds to implicitly co-embed texts embeddings with 3D point clouds embeddings using only scene-level labels. With extensive experiments, we verify that the textual semantic information of category labels is beneficial for 3DSS-VLG which achieves the state-of-the-art on both S3DIS and ScanNet datasets. Further, with an experiment to explore our framework to unobserved scene domains, we demonstrate the generalization capability of our method, which supports its practicality. 

\section*{Acknowledgements}
This work was supported in part by the National Natural Science Foundation of China under Grants 62371310 and 62032015, in part by the Guangdong Basic and Applied Basic Research Foundation under Grant 2023A1515011236, in part by the Stable Support Project of Shenzhen (Project No.20231122122722001), in part by the third phase of high-level university construction of interdisciplinary innovation team project of Shenzhen University(24JCXK03). We also acknowledge the CINECA award under the ISCRA initiative, for the availability of partial HPC resources support, and partial support by the Fundamental Research Funds for the Central Universities, Peking University.

%
%
\bibliographystyle{splncs04}
\bibliography{main}

\begin{thebibliography}{10}
\providecommand{\url}[1]{#1}
\csname url@samestyle\endcsname
\providecommand{\newblock}{\relax}
\providecommand{\bibinfo}[2]{#2}
\providecommand{\BIBentrySTDinterwordspacing}{\spaceskip=0pt\relax}
\providecommand{\BIBentryALTinterwordstretchfactor}{4}
\providecommand{\BIBentryALTinterwordspacing}{\spaceskip=\fontdimen2\font plus
\BIBentryALTinterwordstretchfactor\fontdimen3\font minus \fontdimen4\font\relax}
\providecommand{\BIBforeignlanguage}[2]{{%
\expandafter\ifx\csname l@#1\endcsname\relax
\typeout{** WARNING: IEEEtran.bst: No hyphenation pattern has been}%
\typeout{** loaded for the language `#1'. Using the pattern for}%
\typeout{** the default language instead.}%
\else
\language=\csname l@#1\endcsname
\fi
#2}}
\providecommand{\BIBdecl}{\relax}
\BIBdecl

\bibitem{kweon2022joint}
H.~Kweon and K.-J. Yoon, ``Joint learning of 2d-3d weakly supervised semantic segmentation,'' \emph{Advances in Neural Information Processing Systems}, vol.~35, pp. 30\,499--30\,511, 2022.

\bibitem{yang20232d}
C.-K. Yang, M.-H. Chen, Y.-Y. Chuang, and Y.-Y. Lin, ``2d-3d interlaced transformer for point cloud segmentation with scene-level supervision,'' in \emph{Proceedings of the IEEE/CVF International Conference on Computer Vision.}, 2023, pp. 977--987.

\bibitem{choy20194d}
C.~Choy, J.~Gwak, and S.~Savarese, ``4d spatio-temporal convnets: Minkowski convolutional neural networks,'' in \emph{Proceedings of the IEEE/CVF Conference on Computer Vision and Pattern Recognition}, 2019, pp. 3075--3084.

\bibitem{ghiasi2022scaling}
G.~Ghiasi, X.~Gu, Y.~Cui, and T.-Y. Lin, ``Scaling open-vocabulary image segmentation with image-level labels,'' in \emph{European Conference on Computer Vision}.\hskip 1em plus 0.5em minus 0.4em\relax Springer, 2022, pp. 540--557.

\bibitem{yang2022mil}
C.-K. Yang, J.-J. Wu, K.-S. Chen, Y.-Y. Chuang, and Y.-Y. Lin, ``An mil-derived transformer for weakly supervised point cloud segmentation,'' in \emph{Proceedings of the IEEE/CVF Conference on Computer Vision and Pattern Recognition}, 2022, pp. 11\,830--11\,839.

\bibitem{hu2020randla}
Q.~Hu, B.~Yang, L.~Xie, S.~Rosa, Y.~Guo, Z.~Wang, N.~Trigoni, and A.~Markham, ``Randla-net: Efficient semantic segmentation of large-scale point clouds,'' in \emph{Proceedings of the IEEE/CVF Conference on Computer Vision and Pattern Recognition}, 2020, pp. 11\,108--11\,117.

\bibitem{yan20222dpass}
X.~Yan, J.~Gao, C.~Zheng, C.~Zheng, R.~Zhang, S.~Cui, and Z.~Li, ``2dpass: 2d priors assisted semantic segmentation on lidar point clouds,'' in \emph{European Conference on Computer Vision}.\hskip 1em plus 0.5em minus 0.4em\relax Springer, 2022, pp. 677--695.

\bibitem{qi2017pointnet}
C.~R. Qi, H.~Su, K.~Mo, and L.~J. Guibas, ``Pointnet: Deep learning on point sets for 3d classification and segmentation,'' in \emph{Proceedings of the IEEE/CVF Conference on Computer Vision and Pattern Recognition}, 2017, pp. 652--660.

\bibitem{qi2017pointnet++}
C.~R. Qi, L.~Yi, H.~Su, and L.~J. Guibas, ``Pointnet++: Deep hierarchical feature learning on point sets in a metric space,'' \emph{Advances in neural information processing systems}, vol.~30, 2017.

\bibitem{qian2022pointnext}
G.~Qian, Y.~Li, H.~Peng, J.~Mai, H.~Hammoud, M.~Elhoseiny, and B.~Ghanem, ``Pointnext: Revisiting pointnet++ with improved training and scaling strategies,'' \emph{Advances in Neural Information Processing Systems}, vol.~35, pp. 23\,192--23\,204, 2022.

\bibitem{hegde2023clip}
D.~Hegde, J.~M.~J. Valanarasu, and V.~Patel, ``Clip goes 3d: Leveraging prompt tuning for language grounded 3d recognition,'' in \emph{Proceedings of the IEEE/CVF International Conference on Computer Vision.}, 2023, pp. 2028--2038.

\bibitem{wei2020multi}
J.~Wei, G.~Lin, K.-H. Yap, T.-Y. Hung, and L.~Xie, ``Multi-path region mining for weakly supervised 3d semantic segmentation on point clouds,'' in \emph{Proceedings of the IEEE/CVF Conference on Computer Vision and Pattern Recognition}, 2020, pp. 4384--4393.

\bibitem{ren20213d}
Z.~Ren, I.~Misra, A.~G. Schwing, and R.~Girdhar, ``3d spatial recognition without spatially labeled 3d,'' in \emph{Proceedings of the IEEE/CVF Conference on Computer Vision and Pattern Recognition}, 2021, pp. 13\,204--13\,213.

\bibitem{li2024enhancingrobustnessvisionlanguagemodels}
\BIBentryALTinterwordspacing
J.~Li, Z.~Jie, E.~Ricci, L.~Ma, and N.~Sebe, ``Enhancing robustness of vision-language models through orthogonality learning and cross-regularization,'' 2024. [Online]. Available: \url{https://arxiv.org/abs/2407.08374}
\BIBentrySTDinterwordspacing

\bibitem{hu2021bidirectional}
W.~Hu, H.~Zhao, L.~Jiang, J.~Jia, and T.-T. Wong, ``Bidirectional projection network for cross dimension scene understanding,'' in \emph{Proceedings of the IEEE/CVF Conference on Computer Vision and Pattern Recognition}, 2021, pp. 14\,373--14\,382.

\bibitem{jaritz2019multi}
M.~Jaritz, J.~Gu, and H.~Su, ``Multi-view pointnet for 3d scene understanding,'' in \emph{Proceedings of the IEEE/CVF International Conference on Computer Vision. Workshops}, 2019, pp. 0--0.

\bibitem{liu2022petr}
Y.~Liu, T.~Wang, X.~Zhang, and J.~Sun, ``Petr: Position embedding transformation for multi-view 3d object detection,'' in \emph{European Conference on Computer Vision}.\hskip 1em plus 0.5em minus 0.4em\relax Springer, 2022, pp. 531--548.

\bibitem{zeng2023clip2}
Y.~Zeng, C.~Jiang, J.~Mao, J.~Han, C.~Ye, Q.~Huang, D.-Y. Yeung, Z.~Yang, X.~Liang, and H.~Xu, ``Clip2: Contrastive language-image-point pretraining from real-world point cloud data,'' in \emph{Proceedings of the IEEE/CVF Conference on Computer Vision and Pattern Recognition}, 2023, pp. 15\,244--15\,253.

\bibitem{chen2023clip2scene}
R.~Chen, Y.~Liu, L.~Kong, X.~Zhu, Y.~Ma, Y.~Li, Y.~Hou, Y.~Qiao, and W.~Wang, ``Clip2scene: Towards label-efficient 3d scene understanding by clip,'' in \emph{Proceedings of the IEEE/CVF Conference on Computer Vision and Pattern Recognition}, 2023, pp. 7020--7030.

\bibitem{liang2023open}
F.~Liang, B.~Wu, X.~Dai, K.~Li, Y.~Zhao, H.~Zhang, P.~Zhang, P.~Vajda, and D.~Marculescu, ``Open-vocabulary semantic segmentation with mask-adapted clip,'' in \emph{Proceedings of the IEEE/CVF Conference on Computer Vision and Pattern Recognition}, 2023, pp. 7061--7070.

\bibitem{yun2023ifseg}
S.~Yun, S.~H. Park, P.~H. Seo, and J.~Shin, ``Ifseg: Image-free semantic segmentation via vision-language model,'' in \emph{Proceedings of the IEEE/CVF Conference on Computer Vision and Pattern Recognition}, 2023, pp. 2967--2977.

\bibitem{radford2021learning}
A.~Radford, J.~W. Kim, C.~Hallacy, A.~Ramesh, G.~Goh, S.~Agarwal, G.~Sastry, A.~Askell, P.~Mishkin, J.~Clark \emph{et~al.}, ``Learning transferable visual models from natural language supervision,'' in \emph{International Conference on Machine Learning}, 2021, pp. 8748--8763.

\bibitem{xu2022simple}
M.~Xu, Z.~Zhang, F.~Wei, Y.~Lin, Y.~Cao, H.~Hu, and X.~Bai, ``A simple baseline for open-vocabulary semantic segmentation with pre-trained vision-language model,'' in \emph{European Conference on Computer Vision}.\hskip 1em plus 0.5em minus 0.4em\relax Springer, 2022, pp. 736--753.

\bibitem{xu2023side}
M.~Xu, Z.~Zhang, F.~Wei, H.~Hu, and X.~Bai, ``Side adapter network for open-vocabulary semantic segmentation,'' in \emph{Proceedings of the IEEE/CVF Conference on Computer Vision and Pattern Recognition}, 2023, pp. 2945--2954.

\bibitem{xu2023learning}
J.~Xu, J.~Hou, Y.~Zhang, R.~Feng, Y.~Wang, Y.~Qiao, and W.~Xie, ``Learning open-vocabulary semantic segmentation models from natural language supervision,'' in \emph{Proceedings of the IEEE/CVF Conference on Computer Vision and Pattern Recognition}, 2023, pp. 2935--2944.

\bibitem{chen2023exploring}
J.~Chen, D.~Zhu, G.~Qian, B.~Ghanem, Z.~Yan, C.~Zhu, F.~Xiao, S.~C. Culatana, and M.~Elhoseiny, ``Exploring open-vocabulary semantic segmentation from clip vision encoder distillation only,'' in \emph{Proceedings of the IEEE/CVF International Conference on Computer Vision.}, 2023, pp. 699--710.

\bibitem{bucher2019zero}
M.~Bucher, T.-H. Vu, M.~Cord, and P.~P{\'e}rez, ``Zero-shot semantic segmentation,'' \emph{Advances in Neural Information Processing Systems}, vol.~32, 2019.

\bibitem{xian2019semantic}
Y.~Xian, S.~Choudhury, Y.~He, B.~Schiele, and Z.~Akata, ``Semantic projection network for zero-and few-label semantic segmentation,'' in \emph{Proceedings of the IEEE/CVF Conference on Computer Vision and Pattern Recognition}, 2019, pp. 8256--8265.

\bibitem{hu2022sqn}
Q.~Hu, B.~Yang, G.~Fang, Y.~Guo, A.~Leonardis, N.~Trigoni, and A.~Markham, ``Sqn: Weakly-supervised semantic segmentation of large-scale 3d point clouds,'' in \emph{European Conference on Computer Vision}.\hskip 1em plus 0.5em minus 0.4em\relax Springer, 2022, pp. 600--619.

\bibitem{zhang2022not}
Y.~Zhang, Q.~Hu, G.~Xu, Y.~Ma, J.~Wan, and Y.~Guo, ``Not all points are equal: Learning highly efficient point-based detectors for 3d lidar point clouds,'' in \emph{Proceedings of the IEEE/CVF Conference on Computer Vision and Pattern Recognition}, 2022, pp. 18\,953--18\,962.

\bibitem{chibane2022box2mask}
J.~Chibane, F.~Engelmann, T.~Anh~Tran, and G.~Pons-Moll, ``Box2mask: Weakly supervised 3d semantic instance segmentation using bounding boxes,'' in \emph{European Conference on Computer Vision}.\hskip 1em plus 0.5em minus 0.4em\relax Springer, 2022, pp. 681--699.

\bibitem{genova2021learning}
K.~Genova, X.~Yin, A.~Kundu, C.~Pantofaru, F.~Cole, A.~Sud, B.~Brewington, B.~Shucker, and T.~Funkhouser, ``Learning 3d semantic segmentation with only 2d image supervision,'' in \emph{2021 International Conference on 3D Vision (3DV)}, 2021, pp. 361--372.

\bibitem{alonso20203d}
I.~Alonso, L.~Riazuelo, L.~Montesano, and A.~C. Murillo, ``3d-mininet: Learning a 2d representation from point clouds for fast and efficient 3d lidar semantic segmentation,'' \emph{IEEE Robotics and Automation Letters}, vol.~5, no.~4, pp. 5432--5439, 2020.

\bibitem{cardace2023exploiting}
A.~Cardace, P.~Z. Ramirez, S.~Salti, and L.~Di~Stefano, ``Exploiting the complementarity of 2d and 3d networks to address domain-shift in 3d semantic segmentation,'' in \emph{Proceedings of the IEEE/CVF Conference on Computer Vision and Pattern Recognition}, 2023, pp. 98--109.

\bibitem{ando2023rangevit}
A.~Ando, S.~Gidaris, A.~Bursuc, G.~Puy, A.~Boulch, and R.~Marlet, ``Rangevit: Towards vision transformers for 3d semantic segmentation in autonomous driving,'' in \emph{Proceedings of the IEEE/CVF Conference on Computer Vision and Pattern Recognition}, 2023, pp. 5240--5250.

\bibitem{li2023mseg3d}
J.~Li, H.~Dai, H.~Han, and Y.~Ding, ``Mseg3d: Multi-modal 3d semantic segmentation for autonomous driving,'' in \emph{Proceedings of the IEEE/CVF Conference on Computer Vision and Pattern Recognition}, 2023, pp. 21\,694--21\,704.

\bibitem{hou2021pri3d}
J.~Hou, S.~Xie, B.~Graham, A.~Dai, and M.~Nie{\ss}ner, ``Pri3d: Can 3d priors help 2d representation learning?'' in \emph{Proceedings of the IEEE/CVF International Conference on Computer Vision.}, 2021, pp. 5693--5702.

\bibitem{lahoud20172d}
J.~Lahoud and B.~Ghanem, ``2d-driven 3d object detection in rgb-d images,'' in \emph{Proceedings of the IEEE/CVF International Conference on Computer Vision.}, 2017, pp. 4622--4630.

\bibitem{qi2018frustum}
C.~R. Qi, W.~Liu, C.~Wu, H.~Su, and L.~J. Guibas, ``Frustum pointnets for 3d object detection from rgb-d data,'' in \emph{Proceedings of the IEEE/CVF Conference on Computer Vision and Pattern Recognition}, 2018, pp. 918--927.

\bibitem{xu2018pointfusion}
D.~Xu, D.~Anguelov, and A.~Jain, ``Pointfusion: Deep sensor fusion for 3d bounding box estimation,'' in \emph{Proceedings of the IEEE Conference on Computer Vision and Pattern Recognition}, 2018, pp. 244--253.

\bibitem{zhang2023learning}
R.~Zhang, L.~Wang, Y.~Qiao, P.~Gao, and H.~Li, ``Learning 3d representations from 2d pre-trained models via image-to-point masked autoencoders,'' in \emph{Proceedings of the IEEE/CVF Conference on Computer Vision and Pattern Recognition}, 2023, pp. 21\,769--21\,780.

\bibitem{chen2024towards}
R.~Chen, Y.~Liu, L.~Kong, N.~Chen, X.~Zhu, Y.~Ma, T.~Liu, and W.~Wang, ``Towards label-free scene understanding by vision foundation models,'' \emph{Advances in Neural Information Processing Systems}, vol.~36, 2024.

\bibitem{armeni20163d}
I.~Armeni, O.~Sener, A.~R. Zamir, H.~Jiang, I.~Brilakis, M.~Fischer, and S.~Savarese, ``3d semantic parsing of large-scale indoor spaces,'' in \emph{Proceedings of the IEEE Conference on Computer Vision and Pattern Recognition}, 2016, pp. 1534--1543.

\bibitem{dai2017scannet}
A.~Dai, A.~X. Chang, M.~Savva, M.~Halber, T.~Funkhouser, and M.~Nie{\ss}ner, ``Scannet: Richly-annotated 3d reconstructions of indoor scenes,'' in \emph{Proceedings of the IEEE Conference on Computer Vision and Pattern Recognition}, 2017, pp. 5828--5839.

\bibitem{tatarchenko2018tangent}
M.~Tatarchenko, J.~Park, V.~Koltun, and Q.-Y. Zhou, ``Tangent convolutions for dense prediction in 3d,'' in \emph{Proceedings of the IEEE/CVF Conference on Computer Vision and Pattern Recognition}, 2018, pp. 3887--3896.

\bibitem{shi2019pointrcnn}
S.~Shi, X.~Wang, and H.~Li, ``Pointrcnn: 3d object proposal generation and detection from point cloud,'' in \emph{Proceedings of the IEEE/CVF Conference on Computer Vision and Pattern Recognition}, 2019, pp. 770--779.

\bibitem{zhao2021point}
H.~Zhao, L.~Jiang, J.~Jia, P.~H. Torr, and V.~Koltun, ``Point transformer,'' in \emph{Proceedings of the IEEE/CVF International Conference on Computer Vision.}, 2021, pp. 16\,259--16\,268.

\bibitem{zhang2021weakly}
Y.~Zhang, Z.~Li, Y.~Xie, Y.~Qu, C.~Li, and T.~Mei, ``Weakly supervised semantic segmentation for large-scale point cloud,'' in \emph{Proceedings of the AAAI Conference on Artificial Intelligence}, vol.~35, no.~4, 2021, pp. 3421--3429.

\bibitem{li2022hybridcr}
M.~Li, Y.~Xie, Y.~Shen, B.~Ke, R.~Qiao, B.~Ren, S.~Lin, and L.~Ma, ``Hybridcr: Weakly-supervised 3d point cloud semantic segmentation via hybrid contrastive regularization,'' in \emph{Proceedings of the IEEE/CVF Conference on Computer Vision and Pattern Recognition}, 2022, pp. 14\,930--14\,939.

\bibitem{zhou2016learning}
B.~Zhou, A.~Khosla, A.~Lapedriza, A.~Oliva, and A.~Torralba, ``Learning deep features for discriminative localization,'' in \emph{Proceedings of the IEEE/CVF Conference on Computer Vision and Pattern Recognition}, 2016, pp. 2921--2929.

\bibitem{robert2022learning}
D.~Robert, B.~Vallet, and L.~Landrieu, ``Learning multi-view aggregation in the wild for large-scale 3d semantic segmentation,'' in \emph{Proceedings of the IEEE/CVF Conference on Computer Vision and Pattern Recognition}, 2022, pp. 5575--5584.

\bibitem{xu2023weakly}
X.~Xu, Y.~Yuan, Q.~Zhang, W.~Wu, Z.~Jie, L.~Ma, and X.~Wang, ``Weakly-supervised 3d visual grounding based on visual linguistic alignment,'' \emph{arXiv preprint arXiv:2312.09625}, 2023.

\bibitem{wang2022semaffinet}
Z.~Wang, Y.~Rao, X.~Yu, J.~Zhou, and J.~Lu, ``Semaffinet: Semantic-affine transformation for point cloud segmentation,'' in \emph{Proceedings of the IEEE/CVF Conference on Computer Vision and Pattern Recognition}, 2022, pp. 11\,819--11\,829.

\bibitem{thomas2019kpconv}
H.~Thomas, C.~R. Qi, J.-E. Deschaud, B.~Marcotegui, F.~Goulette, and L.~J. Guibas, ``Kpconv: Flexible and deformable convolution for point clouds,'' in \emph{Proceedings of the IEEE/CVF International Conference on Computer Vision.}, 2019, pp. 6411--6420.

\bibitem{li2022expansion}
J.~Li, Z.~Jie, X.~Wang, X.~Wei, and L.~Ma, ``Expansion and shrinkage of localization for weakly-supervised semantic segmentation,'' in \emph{Advances in Neural Information Processing Systems}, vol.~35, 2022, pp. 16\,037--16\,051.

\bibitem{li2022weakly}
J.~Li, Z.~Jie, X.~Wang, Y.~Zhou, X.~Wei, and L.~Ma, ``Weakly supervised semantic segmentation via progressive patch learning,'' \emph{IEEE Transactions on multimedia}, vol.~25, pp. 1686--1699, 2022.

\bibitem{li2023weakly}
J.~Li, Z.~Jie, X.~Wang, Y.~Zhou, L.~Ma, and J.~Jiang, ``Weakly supervised semantic segmentation via self-supervised destruction learning,'' \emph{Neurocomputing}, vol. 561, p. 126821, 2023.

\end{thebibliography}
\end{document}